\documentclass[11pt,a4paper,dvipsnames]{article}
\usepackage[hyperref]{emnlp2020}
\usepackage{times}
\usepackage{latexsym}

\usepackage{microtype}

\aclfinalcopy 
\def\aclpaperid{1231} 

\usepackage{graphicx}

\usepackage{tabularx}
\usepackage{booktabs}
\usepackage{amsmath}
\usepackage{amsfonts}
\usepackage[nameinlink,capitalise]{cleveref}
\usepackage[inline]{enumitem}
\usepackage{multirow}

\usepackage{tikz}
\usetikzlibrary{shapes}
\usetikzlibrary{positioning}
\usetikzlibrary{arrows.meta}

\newcommand{\para}[1]{\noindent\textbf{#1.}}
\newcommand{\eset}[1]{\left\{ #1 \right\}}
\newcommand{\cset}[2]{\left\{\, #1 \mid #2 \,\right\}}
\newcommand{\supdom}{\mathcal{S}}
\newcommand{\textdom}{\mathcal{T}}
\newcommand{\graphdom}{\mathcal{G}}
\newcommand{\loss}[1]{\mathcal{L}^{\text{#1}}}
\newcommand{\unsup}{\ensuremath{\mathcal{U}}}
\newcommand{\val}{\ensuremath{\mathcal{V}_{100}}}

\newcommand{\noisefont}[1]{\ensuremath{\mathtt{#1}}}
\def\dnrm#1{\mbox{$_{\hbox{\scriptsize #1}}$}}

\newcommand{\ttg}[1]{text#1$\to$#1graph}
\newcommand{\gtt}[1]{graph#1$\to$#1text}
\newcommand{\btg}[1]{text#1$\leftrightarrow$#1graph}

\newcommand*\raiseup[1]{%
	\begingroup
	\setbox0\hbox{\scriptsize\strut #1}%
	\leavevmode
	\raise\dimexpr \ht\strutbox - \ht0\box0
	\endgroup
}
\newcommand{\rttg}{R\raiseup{\ttg{}}}
\newcommand{\rgtt}{R\raiseup{\gtt{}}}

\DeclareMathOperator*{\argmax}{arg\,max}

\DeclareMathOperator*{\expect}{\mathbb{E}}

\title{An Unsupervised Joint System for Text Generation from Knowledge Graphs and Semantic Parsing}

\author{Martin Schmitt\textsuperscript{1}
  \quad Sahand Sharifzadeh\textsuperscript{2}
  \quad Volker Tresp\textsuperscript{2,3}
  \quad Hinrich Sch\"{u}tze\textsuperscript{1}\\\\
  \textsuperscript{1}Center for Information and Language Processing (CIS), LMU Munich\\
  \textsuperscript{2}Department of Informatics, LMU Munich\\
  \textsuperscript{3}Siemens AG Munich\\
  {\tt martin@cis.lmu.de}
}

\date{}

\begin{document}
\maketitle
\begin{abstract}
  Knowledge graphs (KGs)
  can vary greatly from one domain to another.
  Therefore supervised approaches to
  both graph-to-text generation and text-to-graph knowledge extraction (semantic parsing)
  will always suffer
  from a shortage of domain-specific parallel graph-text data;
  at the same time, adapting a model trained on a different domain is often impossible
  due to little or no overlap in entities and relations.
  This situation calls for an approach that
  \begin{enumerate*}[label={(\arabic{*})}]
  	\item does not need large amounts of annotated data and thus
  	\item does not need to rely on domain adaptation techniques to work well in different domains.
  \end{enumerate*}
  To this end, we present the \emph{first approach to
  unsupervised text generation from KGs}
  and show simultaneously how it can be used for \emph{unsupervised semantic parsing}.
  We evaluate our approach on WebNLG v2.1 and a new benchmark leveraging scene graphs from Visual Genome.
  Our system outperforms strong baselines for both \btg{} conversion tasks without any manual adaptation from one dataset to the other.
  In additional experiments, we investigate the impact of using different unsupervised objectives.\footnote{\url{https://github.com/mnschmit/unsupervised-graph-text-conversion}}
\end{abstract}

\section{Introduction}
\label{sec:introduction}

Knowledge graphs (KGs) are a general-purpose approach
for storing information
in a structured, machine-accessible way \citep{harmelen08}.
They are used in various fields and domains to model knowledge about topics as different as
lexical semantics \citep{wordnet05,assem06},
common sense \citep{speer17,sap19},
biomedical research \citep{wishart18} and
visual relations in images \citep{lu16}.

This ubiquity of KGs necessitates interpretability
because diverse users
-- both experts and non-experts --
work with them.
Even though, in principle, a KG is human-interpretable,
non-experts may have difficulty making sense of it.
Thus, there is a need for methods,
such as automatic natural language generation (``\gtt{}''),
that support them.

Semantic parsing,
i.e., the conversion of a text to a formal meaning
representation,
such as a KG, (``\ttg{}'')
is equally important
because
it makes
information that only exists in text form
accessible to machines,
thus assisting knowledge base engineers
in KG creation and completion.

As KGs are so flexible in expressing various kinds of knowledge,
separately created KGs
vary a lot.
This unavoidably leads to a shortage of training data
for both graph$\leftrightarrow$text tasks.
We therefore propose an unsupervised model that
\begin{enumerate*}[label={(\arabic*)}]
\item easily adapts to new KG domains and
\item only requires unlabeled (i.e., non-parallel) texts and graphs from the target domain,
  together with a few fact extraction heuristics,
  but no manual annotation.
\end{enumerate*}

To show the effectiveness of our approach,
we conduct experiments on the latest release (v2.1) of the WebNLG corpus \citep{shimorina-gardent-2018-handling} 
and on a new benchmark we derive from \emph{Visual Genome} \citep{visualgenome}.
While both of these datasets contain enough annotations to train supervised models,
we evaluate our unsupervised approach by ignoring these annotations.
The datasets are particularly well-suited for our evaluation
as both graphs and texts are completely human-generated.
Thus for both our tasks, models are evaluated with natural, i.e., human-generated targets.

Concretely, we make the following contributions:
\begin{enumerate*}[label={(\arabic*)}]
\item We present the first unsupervised non-template approach to text generation from KGs (\gtt{}).
\item We jointly develop a new unsupervised approach to semantic parsing that automatically adjusts to a target KG schema (\ttg{}).
\item In contrast to prior unsupervised \gtt{} and \ttg{} work,
our model does not require manual adaptation to new domains or graph schemas.
\item We provide a thorough analysis of the impact of different unsupervised objectives,
especially the ones we newly introduce for \btg{} conversion.
  \item We create a new large-scale dataset for \btg{} transformation tasks in the visual domain.
\end{enumerate*}

\section{Related Work}
\label{sec:related-work}

\para{\gtt{ }}
Our work is the first attempt at fully unsupervised text generation from KGs.
In this respect it is only comparable to traditional rule- or template-based approaches \citep{kukich83,mcroy00}.
However, in contrast to these approaches,
which need to be manually adapted to new domains and KG schemas,
our method is generally applicable to all kinds of data without modification.

There is a large body of literature about supervised text generation
from structured data, notably about the creation of sports
game summaries from statistical records
\citep{robin95,tanaka98}.  Recent efforts make use of neural
encoder-decoder mechanisms \citep{wiseman17,puduppully19}.
Although text creation from relational databases is related
and our unsupervised method is, in principle, also applicable to this domain,
in our work we specifically address text creation
from graph-like structures such as KGs.

One recent work on supervised text creation from KGs
is \citep{bhowmik18}.
They generate a short description of an entity,
i.e., a single KG node,
based on a set of facts about the entity.
We, however, generate a description of the whole KG,
which involves multiple entities and their relations.
\citet{koncel19} generate texts from whole KGs.
They, however, do not evaluate on human-generated KGs
but automatically generated ones
from the scientific information extraction tool SciIE \citep{luan18}.
Their supervised model is based on message passing
through the topology of the incidence graph of the KG input.
Such graph neural networks \citep{kipf17,velickovic18}
have been widely adopted in supervised graph-to-text tasks \citep{beck18,damonte19,ribeiro19,ribeiro20}.

Even though \citet{marcheggiani-perez-beltrachini-2018-deep} report
that graph neural networks can make better use of graph input than RNNs for supervised learning,
for our unsupervised approach we follow the line of research
that uses RNN-based sequence-to-sequence models \citep{cho-etal-2014-learning,sutskever14}
operating on serialized triple sets \citep{gardent-etal-2017-webnlg,trisedya-etal-2018-gtr,gehrmann-etal-2018-end,castro-ferreira-etal-2019-neural,fan19}.
We make this choice
because learning a common semantic space for both texts and graphs
by means of a shared encoder and decoder
is a central component of our model.
It is a nontrivial, separate research question
whether and how encoder-decoder parameters can effectively be shared
for models working on both sequential and non-sequential data.
We thus leave the adaptation of our approach to graph neural networks for future work.

\para{\ttg{ }}
Converting a text into a KG representation,
our method is an alternative to prior work on open information extraction \citep{niklaus18}
with the advantage that the extractions, though trained without labeled data,
automatically adjust to the KGs used for training.
It is therefore also related to relation extraction in the unsupervised \citep{yao-etal-2011-structured,marcheggiani-titov-2016-discrete,simon-etal-2019-unsupervised}
and distantly supervised setting \citep{riedel10,parikh-etal-2015-grounded}.
However, these systems merely predict a single relation between two given entities in a single sentence,
while we translate a whole text into a KG with potentially multiple facts.

Our \ttg{} task is therefore most closely related to
semantic parsing \citep{kamath19}, but we
convert statements into KG facts whereas semantic parsing
typically converts a question into a KG or database query.
\citet{poon-domingos-2009-unsupervised} proposed the first unsupervised approach.
They, however, still need an additional KG alignment step,
i.e., are not able to directly adjust to the target KG.
Other approaches overcome this limitation but only in exchange for the inflexibility of
manually created domain-specific lexicons \citep{popescu-etal-2004-modern,goldwasser-etal-2011-confidence}.
\citet{poon-2013-grounded}'s approach is more flexible but still relies on preprocessing by a dependency parser,
which generally means that language-specific annotations to train such a parser are needed.
Our approach is end-to-end, i.e., does not need any language-specific preprocessing during inference
and only depends on a POS tagger used in the rule-based \ttg{} system to bootstrap training.

\para{Unsupervised sequence generation}
Our unsupervised training regime for both \btg{} tasks
is inspired by \citep{lample18}.
They used self-supervised pretraining and backtranslation
for unsupervised translation from one language to another.
We adapt these principles and their noise model to our tasks,
and introduce two new noise functions specific to \btg{} conversion.

\section{Preliminaries}
\label{sec:preliminaries}

\subsection{Data structure}

We formalize a KG as a labeled directed multigraph $(V, E, s, t, l)$
where entities are nodes $V$ and
edges $E$ represent relations between entities.
The lookup functions $s, t : E \to V$ assign to each edge
its source and target node.
The labeling function $l$ assigns labels to nodes and edges
where node labels are entity names
and edge labels come from a predefined set $\mathcal{R}$ of relation types.

An equivalent representation of a KG is the set of its
facts. A fact is a triple consisting of an edge's source
node (the subject), the edge itself (the predicate), and its
target node (the object).  So the set of facts $\mathcal{F}$ of
a KG can be obtained from its edges: 
$$\mathcal{F} := \cset{(s(e), e, t(e))}{e \in E}.$$
Applying $l$ to all triple elements
and writing out $\mathcal{F}$ in an arbitrary order
generates a serialization
that makes the KG accessible to sequence models otherwise
used only for text.
This has the advantage that
we can train a sequence encoder to  embed text and
KGs in the same semantic space.
Specifically, we serialize
a KG by writing out its facts separated with end-of-fact
symbols (\texttt{EOF}) and elements of each fact with
special \texttt{SEP} symbols.
We thus define our task as a sequence-to-sequence (seq2seq) task.

\begin{figure}
	\centering
	\includegraphics[width=\linewidth]{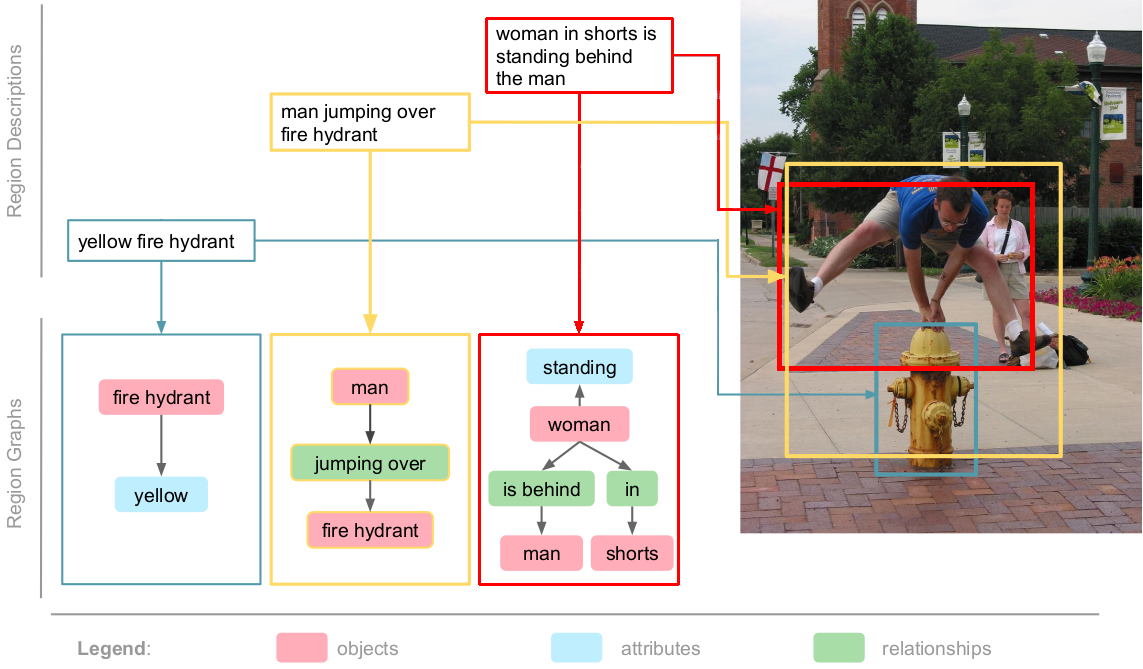}
	\caption{Region graphs and textual region
          descriptions in Visual Genome (VG).
		Image regions serve as common reference for text and graph creation but are disregarded in our work. We solely focus on the pairs of corresponding texts and graphs. Illustration adapted from \citep{visualgenome}.}
	\label{fig:vg_regions}
\end{figure}

\subsection{Scene Graphs}

The Visual Genome (VG) repository is a large collection of
images with associated manually annotated \textbf{scene graphs}; see \cref{fig:vg_regions}.
A scene graph formally describes image objects with their attributes,
e.g., (hydrant, \texttt{attr}, yellow),
and their relations to other image objects,
e.g., (woman, in, shorts).
Each scene graph is organized into smaller
subgraphs, known as \textbf{region graphs},
representing a subpart of a more complex larger picture
that is interesting on its own.
Each region graph is associated with an English text, the \textbf{region description}.
Texts and graphs were not automatically produced from each other,
but were collected from crowdworkers who were
presented an image region and then generated text and graph.
So although the graphs were not specifically
designed to closely resemble the texts,
they describe the same image region.
This semantic correspondence makes scene graph$\leftrightarrow$text
conversion 
an interesting and challenging problem
because text and graph are not simple translations of each other.

\begin{figure}
	\resizebox{\linewidth}{!}{%
		\begin{tikzpicture}
		\node[draw=black,ellipse] (baby) at (0,0) {baby};
		\node[draw=black,ellipse] (blanket) at (-3.5,-1.5) {wrapped in blanket};
		\node[draw=black,ellipse] (small) at (0,-1.5) {small};
		\node[draw=black,ellipse] (hat) at (2,-1.5) {hat};
		\node[draw=black,ellipse] (baseball) at (0,-3) {baseball hat};
		\node[draw=black,ellipse] (pink) at (3,-3) {pink};
		
		\draw[-Latex] (baby) to node[left,yshift=6] {\texttt{attr}} (blanket);
		\draw[-Latex] (baby) to node[left] {\texttt{attr}} (small);
		\draw[-Latex] (baby) to node[right,yshift=3] {wearing} (hat);
		
		\draw[-Latex] (hat) to node[right,yshift=-5,xshift=-4] {\texttt{attr}} (baseball);
		\draw[-Latex] (hat) to node[right] {\texttt{attr}} (pink);
		\end{tikzpicture}
	}
	\caption{Example graph in our new VG benchmark.}
	\label{fig:example_graph}
\end{figure}
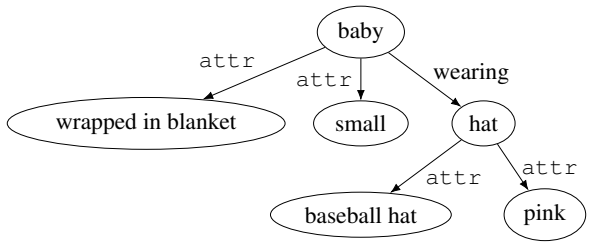

Scene graphs are formalized in the same way as other  KGs:
$V$ here contains image objects and their attributes,
and $\mathcal{R}$ contains all types of visual relationships and the special label \texttt{attr}
for edges between attribute and non-attribute nodes. \cref{fig:example_graph} shows an example.

VG scene graphs have been used before for traditional KG tasks, such as KG completion \citep{wan18},
but we are the first to use them for a \btg{} conversion dataset.

\begin{table*}
	\small
	\centering
	\begin{tabular}{ll}
		\toprule
		noise function & behavior\\
		\midrule
		\noisefont{swap} & applies a random permutation $\sigma$ of words or facts with \\
		& \mbox{$\forall i\in\eset{1,\dots,n}, |\sigma(i) - i| \leq k$}; {$k=3$ for text, $k=+\infty$ for knowledge graphs.} \\
		\midrule
		\noisefont{drop} & removes each fact/word with a probability of $p_{\text{drop}}$.\\
		\midrule
		\noisefont{blank} & replaces each fact/word with a probability of $p_{\text{blank}}$ by a special symbol \texttt{blanked}. \\
		\midrule
		\noisefont{repeat} & inserts repetitions with a probability of $p_{\text{repeat}}$ in a sequence of facts/words.\\
		\midrule
		\noisefont{rule} & generates a noisy translation by applying {\rgtt{}} to a graph or {\rttg{}} to a text. \\
		\bottomrule
	\end{tabular}
	\caption{Noise functions and their behavior on graphs and texts.}
	\label{tab:noise}
\end{table*}

\section{Approaches}
\label{sec:models}

\subsection{Rule-based systems}
\label{sec:rule-based-systems}

We  propose a rule-based system as unsupervised baseline
for each of the \btg{} tasks.
Note that they both assume that the texts are in English.
  
\para{\rgtt{}} 
From a KG serialization,
we remove \texttt{SEP} symbols and
replace \texttt{EOF} symbols by the word \emph{and}.
The special label \texttt{attr} is mapped to \emph{is}.
This corresponds to a template-based enumeration of all KG facts.
See \cref{tab:examples} for an example.

\para{\rttg{}}
After preprocessing a text with NLTK's default POS tagger \citep{nltk}
and removing stop words,
we apply two simple heuristics to extract facts:
\begin{enumerate*}[label={(\arabic*)}]
\item Each verb becomes a predicate; \emph{is} creates facts with predicate \texttt{attr}.
	The content words directly before and after such a predicate word become subject and object.
\item Adjectives $a$ form attributes, i.e., build facts of the form $(X, \texttt{attr}, a)$
  where $X$ is filled with the first noun after $a$.
\end{enumerate*}
These heuristics are similar in nature to a rudimentary parser.
See \cref{tab:ie-qual} for an example.

\begin{figure}
	\centering
	\resizebox{\linewidth}{!}{%
		\begin{tikzpicture}
		\node[align=left,draw=black] (input) {Man wearing a colorful shirt and white pants};
		\node[align=left,below = 2em of input,draw=black] (step1) {\textcolor{red}{Man \texttt{SEP} wearing \texttt{SEP} colorful} \texttt{EOF}\\ shirt \texttt{SEP} \texttt{attr} \texttt{SEP} colorful \texttt{EOF}\\ pants \texttt{SEP} \texttt{attr} \texttt{SEP} white \texttt{EOF}\\ \textcolor{blue}{pants \texttt{SEP} playing \texttt{SEP} tennis} };
		\node[align=left,below = 2em of step1,draw=black] (step2) {pants \texttt{SEP} \texttt{attr} \texttt{SEP} white \texttt{EOF}\\ \textcolor{Orange}{shirt \texttt{SEP} \texttt{attr} \texttt{SEP} colorful} \texttt{EOF}\\ \texttt{blanked} };
		\node[align=left,below = 2em of step2,draw=black] (step3) {pants \texttt{SEP} \texttt{attr} \texttt{SEP} white \texttt{EOF}\\ shirt \texttt{SEP} \texttt{attr} \texttt{SEP} colorful \texttt{EOF}\\ shirt \texttt{SEP} \texttt{attr} \texttt{SEP} colorful \texttt{EOF}\\  \texttt{blanked} };
		
		\draw[-Latex] (input) to node[right] {\noisefont{rule}} (step1);
		\draw[-Latex] (step1) to node[right] {$\noisefont{blank}\circ\noisefont{drop}\circ\noisefont{swap}$} (step2);
		\draw[-Latex] (step2) to node[right] {\noisefont{repeat}} (step3);
		
		\draw[-Latex,bend left=90] (step3) to node[right,] {\Large $\loss{lm}$} (input);
		\end{tikzpicture}
	}
	\caption{Example noisy training instance for the \gtt{} task in the composed noise setting. The fact highlighted in red is removed by \noisefont{drop}, the one in blue is replaced with \texttt{blanked} by \noisefont{blank}, the one in orange is repeated by \noisefont{repeat}.}
	\label{fig:noise_example}
\end{figure}
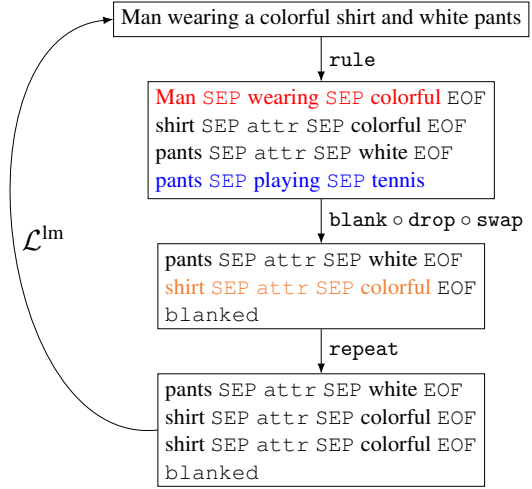

\subsection{Neural seq2seq systems}
\label{sec:neur-transl-syst}

Our main system is a neural seq2seq architecture.
We equip the standard encoder-decoder model with attention \citep{bahdanau14}
and copy mechanism \citep{gu16}.
Allowing the model to directly copy from the source to the target side
is beneficial in data to text generation  \citep{wiseman17,puduppully19}.
The encoder (resp.\ decoder) is a 
bidirectional (resp.\ unidirectional) LSTM \citep{lstm}.
Dropout \citep{dropout} is applied
at the input of both encoder and decoder
\citep{britz17}.
We combine this model with the following concepts:

\para{Multi-task model}
In unsupervised machine translation,
systems are trained for both translation directions
\citep{lample18}. In the same way,
we train our system for both conversion tasks \btg{},
sharing encoder and decoder.
To tell the decoder which type of output should be produced (text or graph),
we initialize the cell state of the  decoder
with an embedding of the desired
output type.
The hidden state of the  decoder is initialized
with the last state of the encoder as usual.

\para{Noisy source samples}
\citet{lample18-noise} introduced denoising auto-encoding
as pretraining and auxiliary task
to train the decoder to produce well-formed output
and make the encoder robust to noisy input.
The training examples for this task consist of
a noisy version of a sentence as source
and the original sentence as target.
We adapt this idea and propose the following noise functions
for the domains of graphs and texts:
\noisefont{swap}, \noisefont{drop}, \noisefont{blank}, \noisefont{repeat}, \noisefont{rule}.
\cref{tab:noise} describes their behavior.
\noisefont{swap}, \noisefont{drop} and \noisefont{blank} are adapted from \citep{lample18-noise}
with facts in graphs taking the role of words in text.
As order should be irrelevant in a set of facts,
we drop the locality constraint
in the \noisefont{swap} permutation
for graphs by setting $k=+\infty$.

Denoising samples
generated by \noisefont{repeat}
requires the model to learn to
remove redundant information in a set of facts.
In the
case of text, \noisefont{repeat}
mimics a behavior
often observed with insufficiently trained neural models,
i.e., repeating words considered important.

Unlike the other noise functions,
\noisefont{rule} does not ``perturb'' its input,
but rather noisily backtranslates it.
We will see in \cref{sec:ablation-study}
that bootstrapping with these noisy translations
is essential.

We consider two fundamentally different noise injection regimes:
\begin{enumerate*}[label={(\arabic{*})}]
\item The \textbf{composed noise} setting is an adaptation of \citet{lample18-noise}'s
noise model ($\noisefont{blank}\circ\noisefont{drop}\circ\noisefont{swap}$)
where our newly introduced noise functions \noisefont{rule} and \noisefont{repeat}
are added to the start and end of the pipeline, i.e.,
all data samples are treated equally with the same noise function
$C_\text{comp} := \noisefont{repeat}\circ\noisefont{blank}\circ\noisefont{drop}\circ\noisefont{swap}\circ\noisefont{rule}$.
\Cref{fig:noise_example} shows an example.
\item In the \textbf{sampled noise} setting,
we do not use all noise functions at once but sample a single one per data instance.
\end{enumerate*}

\subsection{Training regimes}
\label{sec:training-regimes}

We denote the sets of graphs and corresponding texts
by $\graphdom$ and $\textdom$.
The set of available supervised examples $(x, y) \in \graphdom\times\textdom$
is called $\supdom \subset \graphdom\times\textdom$.
$P_{g}$ and $P_{t}$ are
probabilistic
models that
generate,
conditioned on
any input,
a graph ($g$) or a text ($t$).

\para{Unsupervised training}
We first obtain a language model for both graphs and text
by training one epoch with the denoising auto-encoder objective:
\begin{align}
  \loss{denoise} &= \expect_{x\sim\graphdom}[-\operatorname{log} P_{g}(x|C(x))] +{} \notag \\
            &\phantom{{}={}}
              \expect_{y\sim\textdom}[-\operatorname{log}
              P_{t}(y|C(y))]  \nonumber
\end{align}
where $C \in \eset{C_\text{comp}}$ for composed noise
and
$C\in\eset{\noisefont{swap}, \noisefont{blank}, \noisefont{drop}, \noisefont{repeat}, \noisefont{rule}}$
 for sampled noise.
In this pretraining epoch only,
we use all possible noise functions individually on all
available data.
As sampled noise incorporates five different noise functions and composed noise only one,
this results in five times more pretraining samples
for sampled noise than for composed noise.

In subsequent epochs,
we additionally consider $\loss{back}$ as training signal:
\begin{align}
    \loss{back} &= \expect_{x\sim\graphdom}[-\operatorname{log} P_{g}(x|z^*(x))] +{} \notag \\
                &\phantom{{}={}}
    \expect_{y\sim\textdom}[-\operatorname{log}
      P_{t}(y|w^*(y))] \nonumber \\
  z^*(x) &= \argmax_z P_{t}(z|x)\nonumber \\
  w^*(y) &= \argmax_w P_{g}(w|y)\nonumber
\end{align}
This means that, in each iteration, we apply the current model
to backtranslate a text (graph) to obtain a
potentially imperfect graph (text) that we can use as noisy
source with the clean original input being the target.
This gives us
a pseudo-parallel training instance for the next
iteration -- recall that we address unsupervised generation,
i.e., without access
to parallel data.

The total loss in these epochs is $\loss{back} + \loss{denoise}$,
where now $\loss{denoise}$ only samples one possible type of noise
independently for each data instance. 

\para{Supervised training}
Our intended application is an unsupervised scenario.
For our two datasets, however,
we have labeled data (i.e., a ``parallel corpus'') and
so can also compare our model to its supervised variant.
Although supervised
performance is generally better,
it serves as a reference point and
gives us an idea of the impact of supervision
as opposed to factors like model architecture and hyperparameters.
The supervised loss is simply defined as follows:
\begin{align}  
  \loss{sup} &= \expect_{(x, y)\sim\supdom}
  ~[-\operatorname{log} P_{t}(y|x) - \operatorname{log}
    P_{g}(x|y)]  \nonumber
\end{align}

\begin{table}
	\centering
	\small
	\def\colsep{\hspace{.75em}}
	\begin{tabularx}{\linewidth}{X@{\hspace{.3em}}r@{\colsep}r@{\colsep}r}
		\toprule
		& VG & VG\dnrm{ball} & WebNLG \\
		\midrule
		train split size & 2,412,253 & 151,790 & 34,338 \\
		val split size & 323,478 & 21,541 & 4,313 \\
		test split size & 324,664 & 20,569 & 4,222 \\
		\midrule
		\#{}relation types & 36,506 & 5,167 & 373 \\
		avg \#{}facts in graph & 2.7 & 2.5 & 3.0 \\
		avg \#{}tokens in text & 5.4 & 5.5 & 22.8 \\
		\midrule
		avg\,\%{}\,text\,tokens\,in\,graph & 49.3 & 50.6 & 49.4 \\
		avg\,\%{}\,graph\,tokens\,in\,text & 52.3 & 54.7 & 75.6 \\
		\bottomrule
	\end{tabularx}
	\caption{Statistics of WebNLG v2.1 and our newly created benchmark VG;
		VG\dnrm{ball} is a subset of VG representing images from ball sports events. Data split sizes are given as number of graph-text pairs.}
	\label{tab:datastats}
\end{table}

\begin{table*}
	\footnotesize
	\centering
	\def\columnwidth{\hspace{.8em}}
	\begin{tabular}{lrrrrrrrrrrrr}
		\toprule
		& \multicolumn{6}{c}{Visual Genome} & \multicolumn{6}{c}{WebNLG}\\
		\cmidrule(lr){2-7}\cmidrule(lr){8-13}
		\gtt{ }&\multicolumn{2}{c}{BLEU} & \multicolumn{2}{c}{METEOR} & \multicolumn{2}{c}{CHRF++} & \multicolumn{2}{c}{BLEU} & \multicolumn{2}{c}{METEOR} & \multicolumn{2}{c}{CHRF++}\\
		\cmidrule(lr){2-3}\cmidrule(lr){4-5}\cmidrule(lr){6-7}\cmidrule(lr){8-9}\cmidrule(lr){10-11}\cmidrule(lr){12-13}
		& val & test & val & test & val & test & val & test& val & test& val & test\\
		\midrule
		\rgtt{} & 5.9 & 5.9 & 28.2 & 28.1 & 43.4 & 43.3 & 18.3 & 18.3 & 33.5 & 33.6 & 55.0 & 55.2 \\
		Ours w/ sampled noise & 19.8 & 19.5 & 31.4 & 31.2 & 50.9 & 50.7 & \textbf{39.1} & \textbf{37.7} & \textbf{35.4} & \textbf{35.5} & \textbf{61.9} & \textbf{62.1} \\
		Ours w/ composed noise & \textbf{23.2} & \textbf{23.2} & \textbf{33.0} & \textbf{32.9} & \textbf{53.7} & \textbf{53.6} & 30.8 & 30.5 & 30.2 & 30.0 & 53.1 & 52.8 \\
		\midrule
		Ours \emph{supervised} & 26.5 & 26.4 & 32.3 & 32.2 & 53.7 & 53.6 & 35.1 & 34.4 & 39.6 & 39.5 & 64.1 & 64.0 \\
		\bottomrule
	\end{tabular}
	\caption{Results for unsupervised and supervised
          text generation.
          Note that training a supervised model on millions of labeled samples is usually not an option. Best unsupervised models are identified by best BLEU on \val{}. BLEU and METEOR are computed with scripts from \citep{lin18}; the CHRF++ script is from \citep{popovic17}.}
	\label{tab:gtt}
\end{table*}

\section{Experiments}
\label{sec:experiments}

\subsection{Data}
\label{sec:experiment-data}

For our experiments,
we randomly split the VG images
80/10/10 into train/val/test.
We then
remove all graphs from train
that also occur in one of the images in val or test.
Finally,
we unify graph serialization duplicates with different texts
to single instances with multiple references for \gtt{}
and proceed analogously with text duplicates for \ttg{}.
For WebNLG v2.1, we use the data splits as provided.
Following \citep{gardent-etal-2017-creating},
we resolve the camel case of relation names and remove underscores from entity names
in a preprocessing step.
For both datasets, the order of facts in graph serializations corresponds to the order of triples in the original dataset.
Because of VG's enormous size and limited computation power,
we additionally create a closed-domain ball sports subset of VG, called VG\dnrm{ball},
which we can use to quickly conduct additional experiments (see \cref{sec:ablation-study}).
We identify all images
where at least one region graph contains at least one fact
that mentions an object ending with \emph{ball}
and take all  regions from them
(keeping data splits the same).
In contrast to alternatives like random subsampling,
we consider this domain-focused construction
more realistic.

\cref{tab:datastats} shows relevant statistics for all datasets.
While VG and WebNLG have similar statistics,
VG is around 70 times larger than WebNLG,
which makes it an interesting benchmark for future
research,
both supervised and unsupervised.
Apart from size, there are two important differences:
\begin{enumerate*}[label={(\arabic{*})}]
	\item The VG graph schema has been freely defined by crowd workers
	and thus features a large variety of different relations.
	\item The percentage of graph tokens occurring in the text,
	a measure important for the \ttg{} task,
	is lower for VG than for WebNLG.
Thus, VG graphs contain more details than their corresponding texts,
	which is a characteristic feature of the domain of image captions:
	they mainly describe the salient image parts.
\end{enumerate*}

\subsection{Training details}
\label{sec:training-details}

We train all models
with the Adam optimizer \citep{adam}
for maximally 30 epochs.
We stop supervised models early
when $\loss{sup}$ does not decrease on val for 10 epochs.
Unsupervised models
are stopped after 5 iterations on VG
because of its big size and limited computational resources.
All hyperparameters and more details are described in \cref{app:hparams,app:model_details}.
Our implementation is based on AllenNLP \citep{allennlp}.

In unsupervised training,
input graphs and texts are the same
as in supervised training --
only the gold target sides are ignored.
While it is an artificial setup to split paired data and treat them as unpaired,
this not only makes the supervised and unsupervised settings more directly comparable,
but also ensures that the text data resemble the evaluation texts in style and domain.
For the purpose of experiments on a benchmark, this seems appropriate to us.
For a concrete use case, it would be an important first step to find adequate texts
that showcase the desired language style and that are about a similar topic as the KGs
that are to be textualized.
As KGs are rarely the only means of storing information, e.g., in an industrial context,
such texts should not be hard to come by in practice.

\section{Results and Discussion}
\label{sec:results}

\subsection{Text generation from graphs}

\begin{table}[t]
	\footnotesize
	\centering
	\setlength\tabcolsep{.5em}
	\begin{tabular}{l|rrrr|rrrr}
		\toprule
		\multicolumn{1}{c}{}& \multicolumn{4}{c}{sampled noise} & \multicolumn{4}{c}{composed noise}\\
		\cmidrule(lr){2-5}\cmidrule(lr){6-9}
		\multicolumn{1}{c}{\#{}}& \unsup & \val & val & \multicolumn{1}{r}{test} & \unsup & \val & val & test \\
		\midrule
		1 & \textbf{80.4} & 7.8 & 10.1 & 9.9 & \textbf{72.2} & 15.9 & 19.8 & 19.7 \\
		2 & 50.7 & 7.2 & 9.2 & 9.1 & 41.2 & 14.0 & 15.2 & 15.1 \\
		3 & 67.6 & 19.5 & 19.4 & 19.2 & 61.0 & 22.7 & \textbf{23.5} & \textbf{23.4} \\
		4 & 56.4 & \textbf{21.2} & \textbf{19.8} & \textbf{19.5} & 51.9  & 22.2 & 21.4 & 21.3 \\
		5 & 62.9 & 20.0 & 19.6 & 19.4 & 60.5 & \textbf{24.5} & 23.2 & 23.2 \\
		\bottomrule
	\end{tabular}
	\caption{BLEU scores on VG for our unsupervised models evaluated for \gtt{} at different
		iterations. \unsup{} is calculated on all unlabeled data used for training.
		\val{} is a 100-size random sample from val. All results are computed with scripts from \citep{lin18}.}
	\label{tab:model-select}
\end{table}

\para{Model selection}
\Cref{tab:model-select} shows how performance of our unsupervised model changes at every backtranslation iteration,
measured in BLEU \citep{bleu},
a common metric for natural language generation.
For model selection,
we adopt the two methods proposed by \citet{lample18},
i.e.,
a small validation set
(we take a 100-size random subset of val, called \val)
and a fully unsupervised criterion (\unsup)
where BLEU compares an unlabeled sample with its back-and-forth translation.
We confirm their finding
that \unsup{} is not reliable for neural text generation models
whereas \val{} correlates better with performance on the larger test sets.
We use \val{} for model selection in the rest of this paper.

\para{Quantitative evaluation}
\Cref{tab:gtt} shows BLEU, METEOR \citep{meteor} and CHRF++ \citep{popovic-2017-chrf}
for our unsupervised models and the rule baseline \rgtt{},
which is in many cases, i.e., if parallel graph-text data are scarce, the only alternative.

First, we observe that \rgtt{} performs much better on WebNLG than VG,
indicating that our new benchmark poses a tougher challenge.
Second, our unsupervised models consistently outperform this baseline on all metrics and on both datasets,
showing that our method produces textual descriptions
much closer to human-generated ones.
Third,
noise composition,
the general default in unsupervised machine translation,
does not always perform better than noise sampling.
Thus, it is worthwhile to try different noise settings for new tasks or datasets.

Surprisingly,
supervised and unsupervised models perform nearly on par.
Real supervision does not seem to give much better guidance in training than our unsupervised regime,
as measured by our three metrics on two different datasets.
Some metric-dataset combinations even favor one of the unsupervised models.
Our qualitative observations provide a possible explanation for that.

\begin{table}[t]
	\centering
	\small
	\begin{tabularx}{\linewidth}{@{}c@{\hspace{.3em}}lX@{}}
		\toprule
		(a) &Reference text & a baseball cap on a baby's head \\
		\midrule
		(b) &\rgtt{} & baby is small and baby is wrapped in blanket and hat is pink and hat is baseball hat and baby wearing hat \\
		\midrule
		(c) &Unsuperv.\ neural & small baby wrapped in blanket\\
		&model& with pink baseball hat\\
		\midrule
		(d) &Superv.\ neural model & baby wearing a pink hat\\
		\bottomrule
	\end{tabularx}
	\caption{Texts generated from graph in \cref{fig:example_graph}.}
	\label{tab:examples}
\end{table}

\para{Qualitative observations}
Taking a look at example generations (\cref{tab:examples}),
we also see qualitatively how much easier it is
to grasp the content of our natural language summarization
than reading through a simple enumeration of KG facts.
We find that the unsupervised model (c) seems to output the KG information in a more complete manner
than its supervised counterpart (d).
The supervision probably introduces a bias present in the training data
that image captions focus on salient image parts
and therefore the supervised model is encouraged to omit information.
As it never sees a corresponding text-graph pair together,
the unsupervised model cannot draw such a conclusion.

\begin{table}[t]
	\footnotesize
	\centering
	\def\columnwidth{\hspace{.8em}}
	\begin{tabular}{l|r@{\columnwidth}r@{\columnwidth}r@{\columnwidth}r|r@{\columnwidth}r@{\columnwidth}r@{\columnwidth}r}
		\toprule
		\multicolumn{1}{c}{}&\multicolumn{4}{c}{sampled noise}&\multicolumn{4}{c}{composed noise}\\
		\cmidrule(lr){2-5}\cmidrule(lr){6-9}
		\multicolumn{1}{c}{\#{}}& \unsup & \val & val & \multicolumn{1}{r}{test} & \unsup & \val & val & \multicolumn{1}{r}{test} \\
		\midrule
		1 & 19.1 & 1.0 & 1.2 & 1.2 & 17.0 & 2.0 & 2.2 & 2.2\\
		2 & \textbf{71.0} & \textbf{21.7} & \textbf{19.1} & \textbf{18.8} & 49.3 & \textbf{22.1} & \textbf{22.1} & \textbf{21.7} \\
		3 & 58.2 & 19.3 & 18.6 & 18.3 & 45.9 & 18.7 & 19.7 & 19.4 \\
		4 & 62.3 & 18.3 & \textbf{19.1} & \textbf{18.8} & \textbf{54.4} & 19.9 & 20.8 & 20.5 \\
		5 & 63.7 & 19.8 & 19.0 & 18.7 & 49.0 & 18.8 & 19.0 & 18.8 \\
		\bottomrule
	\end{tabular}
	\caption{F1 scores on VG for our models from \cref{tab:model-select} evaluated on \ttg{} at different iterations.}
	\label{tab:unsup:text2graph}
\end{table}

\begin{table}[t]
	\footnotesize
	\centering
	\def\bigstrutjot{\dimexpr \aboverulesep + \belowrulesep + \cmidrulewidth}
	\begin{tabularx}{\linewidth}{Xrrrr}
		\toprule
		\multirow{2}[1]{*}{\ttg{ }}& \multicolumn{2}{c}{VG} & \multicolumn{2}{c}{WebNLG} \\
		\cmidrule(lr){2-3}\cmidrule(lr){4-5}
		&val & test &val & test \\
		\midrule
		\rttg{} & 13.4 & 13.1 & 0.0 & 0.0 \\
		Stanford SG Parser & 19.5 & 19.3 & 0.0 & 0.0 \\
		Ours w/ sampled noise & 19.1 & 18.8 & \textbf{38.5} & \textbf{39.1} \\
		Ours w/ composed noise & \textbf{22.1} & \textbf{21.7} & 32.5 & 33.1 \\
		\midrule
		Ours \emph{supervised} & 23.5 & 23.0 & 52.8 & 52.8 \\
		\bottomrule
	\end{tabularx}
	\caption{F1 scores of facts extracted by our unsupervised semantic parsing (\ttg{}) systems and 
		our model trained with supervision.}
	\label{tab:ttg}
\end{table}

\begin{table}[t]
	\centering
	\small
	\begin{tabularx}{1.0\linewidth}{@{\hspace{.4em}}l@{\hspace{.7em}}X@{\hspace{.4em}}}
		\toprule
		Input sentence & Man wearing a colorful shirt and white pants playing tennis\\
		\midrule
		Reference (RG) & \mbox{(shirt, \texttt{attr}, colorful)} \mbox{(pants, \texttt{attr}, white)} \mbox{(man, wearing, shirt)} \mbox{(man, wearing, pants)} \\
		\midrule
		\rttg{} & \colorbox{red}{(Man, wearing, colorful)} \colorbox{green}{(shirt, \texttt{attr}, colorful)} \colorbox{green}{(pants, \texttt{attr}, white)} \colorbox{red}{(pants, playing, tennis)}\\
		\midrule
		Stanford Scene & \colorbox{red}{(shirt, play, tennis)}, \\
		Graph Parser   & \colorbox{red}{(pants, play, tennis)}, \\
		& \colorbox{green}{(shirt, \texttt{attr}, colorful)},\\
		& \colorbox{green}{(pants, \texttt{attr}, white)} \\
		\midrule
		Unsuperv.\ model & \colorbox{red}{(pants, \texttt{attr}, colorful)}\\
		w/ composed noise & \colorbox{green}{(pants, \texttt{attr}, white)}\\
		& \colorbox{green}{(man, wearing, shirt)} \colorbox{yellow}{(man, playing, tennis)}\\
		\midrule
		Supervised model & \colorbox{green}{(shirt, \texttt{attr}, colorful)} \\
		& \colorbox{green}{(pants, \texttt{attr}, white)} \colorbox{green}{(Man, wearing, shirt)} \colorbox{green}{(Man, wearing, pants)}\\
		\bottomrule
	\end{tabularx}
	\caption{Example fact extractions and evaluation wrt
		reference graph (RG).
		Green: correct ($\in$ RG). Yellow: acceptable fact, but
		$\notin$ RG. Red: incorrect ($\notin$ RG).}
	\label{tab:ie-qual}
\end{table}

\subsection{Graph extraction from texts}

We evaluate semantic parsing (\ttg{}) performance
by computing the micro-averaged F1 score of extracted facts.
If there are multiple reference graphs (cf. \cref{sec:experiment-data}),
an extracted fact is considered correct
if it occurs in at least one reference graph.
For the ground truth number of facts to be extracted from a given text,
we take the maximum number of facts of all its reference graphs.

\para{Model selection}
\cref{tab:unsup:text2graph} shows
that
(compared to text generation quality)
\unsup{} is more reliable for \ttg{} performance.
For sampled noise, it correctly identifies the best iteration,
whereas for composed noise it chooses second best.
In both noise settings,
\val{} perfectly chooses the best model.

\para{Quantitative observations}
\Cref{tab:ttg} shows a comparison of our unsupervised models
with two rule-based systems,
our \rttg{} and
the highly domain-specific Stanford Scene Graph Parser (SSGP) by \citet{sg_parser}.

We choose these two baselines to adequately represent the state of the art in the unsupervised setting.
Recall from \cref{sec:related-work} that the only previous unsupervised works either cannot adapt to a target graph schema (open information extraction), which means their precision and recall of retrieved facts is always 0, or have been created for SQL query generation from natural language questions \citep{poon-2013-grounded}, a related task that is yet so different that an adaptation to triple set generation from natural language statements is nontrivial.
While rule-based systems do not automatically adapt to new graph schemas either,
\rttg{} and SSGP were at least designed with the scene graph domain in mind.

Although SSGP was not optimized to match the scene graphs from VG,
its rules were still engineered to cover typical idiosyncrasies of textual image descriptions and corresponding scene graphs.
Besides, we evaluate it with lemmatized reference graphs
because it only predicts lemmata as predicates.
All this gives it a major advantage over the other presented systems
but it is nonetheless outperformed by our best unsupervised model
-- even on VG.
This shows that our automatic method can beat even hand-crafted domain-specific rules.

Both \rttg{} and SSGP fail to predict any fact from WebNLG.
The DBpedia facts from WebNLG often contain multi-token entities
while \rttg{} only picks single tokens from the text.
Likewise, SSGP models multi-token entities as two nodes with an \texttt{attr} relation.
This illustrates the importance of automatic adaptation to the target KG.
Although our system uses \rttg{} during unsupervised training
and is similarly not adapted to the WebNLG dataset,
it performs significantly better.

Supervision helps more on WebNLG than on VG.
The poor performance of \rttg{} on WebNLG is probably a handicap for unsupervised learning.

\para{Qualitative observations}
\cref{tab:ie-qual} shows example facts extracted by different systems.
\rttg{} and SSGP are both fooled by the proximity of the noun \emph{pants} and the verb \emph{play}
whereas our model correctly identifies \emph{man} as the subject.
It, however, fails to identify \emph{shirt} as an
entity and associates the two attributes \emph{colorful} and \emph{white} to \emph{pants}.
Only the supervised model produces perfect output.

\begin{table}[t]
	\footnotesize
	\centering
	\begin{tabularx}{\linewidth}{Xrrrr}
		\toprule
		& \multicolumn{2}{c}{VG\dnrm{ball}} & \multicolumn{2}{c}{WebNLG} \\
		\cmidrule(lr){2-3}\cmidrule(lr){4-5}
		& g$\to$t & t$\to$g & g$\to$t & t$\to$g \\
		& BLEU & F1 & BLEU & F1 \\
		\midrule
		No noise & \underline{0.9} & \underline{0.0} & \underline{14.8} & \underline{0.0} \\
		sample all noise funs & \textbf{19.9} & 17.3 & \textbf{39.1} & \textbf{38.5} \\
		compose all noise funs & 19.6 & \textbf{19.0} & 30.8 & 32.5 \\
		\midrule
		use only \noisefont{rule}  & \textbf{19.5} &  \textbf{18.5} & 37.4 & \textbf{31.0} \\
		use only \noisefont{swap}  & \underline{0.9} & \underline{0.0} & \underline{13.1} & \underline{0.0} \\
		use only \noisefont{drop}  & \underline{0.9} & \underline{0.0} & \textbf{39.9} & 30.1 \\
		use only \noisefont{blank} & \underline{0.9} & \underline{0.0} & \underline{14.9} & \underline{0.0} \\
		use only \noisefont{repeat}& \underline{1.1} & \underline{0.0} & \underline{15.7} & \underline{0.0} \\
		\midrule
		sample all but \noisefont{rule} & \underline{0.9} & \underline{0.0} & \underline{14.9} & \underline{0.0} \\
		sample all but \noisefont{swap} & 19.2 & 17.0 & 39.6 & \textbf{37.3} \\
		sample all but \noisefont{drop} & 19.5 & 16.0 & 39.2 & 35.3 \\
		sample all but \noisefont{blank} & 19.9 & \textbf{17.5} & \textbf{41.0} & 37.0 \\
		sample all but \noisefont{repeat} &\textbf{20.4} & 16.6 & 36.7 & 37.1 \\
		\midrule
		comp.\ all but \noisefont{rule} & \underline{0.9} & \underline{0.0} & \underline{13.5} & \underline{0.0} \\
		comp.\ all but \noisefont{swap} & 20.2 & 16.3 & 35.9 & 40.8 \\
		comp.\ all but \noisefont{drop} & 21.5 & 18.6 & 36.4 & 41.1 \\
		comp.\ all but \noisefont{blank} & 20.2 & 16.3 & 34.8 & 40.4 \\
		comp.\ all but \noisefont{repeat} & \textbf{21.1} & \textbf{20.1} & \textbf{38.5} & \textbf{42.3} \\
		\bottomrule
	\end{tabularx}
	\caption{Ablation study of our models on val of
		VG\dnrm{ball} and WebNLG v2.1. 
		Models selected based on $\val$.
		Bold: best performance per column and block.
		Underlined: worse than corresponding rule-based system.}
	\label{tab:ablation}
\end{table}

\subsection{Noise and translation completeness}

Sampled noise only creates training pairs that either are complete rule-based translations or reconstruction pairs from a noisy graph to a complete graph or a noisy text to a complete text.
In contrast, composed noise can introduce translations from a noisy text to a complete graph or vice versa
and thus encourage a system to omit input information (cf.\ \cref{fig:noise_example}).
This difference is mirrored nicely in the results of our unsupervised systems for both tasks:
composed noise performs better on VG where omitted information in an image caption is common
and sampled noise works better on WebNLG where the texts describe their graphs completely.

\section{Noise Ablation Study}
\label{sec:ablation-study}

Our unsupervised objectives are defined by different types of noise models.
Hence, we examine their impact in a noise ablation study.
\cref{tab:ablation} shows results
for \ttg{} and \gtt{}
on the validation splits of VG\dnrm{ball} and WebNLG.

For both datasets and tasks,
introducing variation via noise functions is
crucial for the success of unsupervised learning.
The model without noise (i.e., $C(x) = x$)
fails completely
as do all models lacking \noisefont{rule} as type of noise,
the only exception being the only-\noisefont{drop} system on WebNLG.
Even though \noisefont{drop} seems to work equally well in this one case,
the simple translations
delivered by our rule-based systems
clearly provide the most useful information for the unsupervised models
-- notably in combination with the other noise functions:
removing \noisefont{rule} and keeping all other types of noise
(cf.\ ``sample all but \noisefont{rule}'' and ``comp.\ all but \noisefont{rule}'')
performs much worse than leaving out \noisefont{drop}.

We hypothesize that our two rule systems provide two important pieces of information:
\begin{enumerate*}[label=(\arabic*)]
	\item \rgtt{} helps distinguish data format tokens from text tokens and
	\item \rttg{} helps find probable candidate words in a text that form facts for the data output.
\end{enumerate*}
As opposed to machine translation,
where usually every word in a sentence is translated into a fluent sentence in the target language,
identifying words that probably form a fact is more important in data-to/from-text generation.

We moreover observe that
our unsupervised models always
improve on the rule-based systems
even when \noisefont{rule} is the only type of noise:
\gtt{} BLEU increases from 6.2/18.3 to 19.5/37.4 on VG\dnrm{ball}/WebNLG
and \ttg{} F1 from 14.4/0.0 to 18.5/31.0.

Finally,
our ablation study makes clear that
there is no best noise model for all datasets and tasks.
We therefore recommend experimenting with
both different sets of noise functions and noise injection regimes (sampled vs.\ composed)
for new data.

\section{Conclusion}
\label{sec:conclusion}

We presented the first fully unsupervised approach to text generation from KGs
and a novel approach to unsupervised semantic parsing that automatically adapts to a target KG.
We showed the effectiveness of our approach on two datasets,
WebNLG v2.1 and a new \btg{} benchmark in the visual domain,
derived from Visual Genome.
We  quantitatively and qualitatively analyzed our method
on \btg{} conversion.
We explored the impact of different unsupervised objectives
in an ablation study and
found that
our newly introduced unsupervised objective using rule-based translations
is essential for the success of unsupervised learning.

\section*{Acknowledgments}

We thank the anonymous reviewers for their helpful comments
and gratefully acknowledge a Ph.D.\ scholarship awarded to the first author
by the German Academic Scholarship Foundation (Studienstiftung des deutschen Volkes).
This work was supported by by the BMBF as part of the project MLWin (01IS18050).

\bibliographystyle{acl_natbib}
\bibliography{references,anthology}

\appendix
\section{Hyperparameters}
\label{app:hparams}
We use the following settings for all our experiments:
learning rate of $10^{-4}$, word embeddings of size 300, an LSTM hidden size of 250, a dropout rate of 0.2 and a batch size of 10.
Following \citet{lample18},
we set $p_{\text{blank}} = p_{\text{repeat}} = 0.2$, $p_{\text{drop}} = 0.1$. 
For inference, we decode greedily with a maximum number of 40 decoding steps.
To speed up unsupervised learning, we increase the batch size to 64 when creating backtranslations.

\section{Model details}
\label{app:model_details}
We train with homogeneous batches of one target output type (text or graph) at a time.
We use a single GeForce GTX 1080 GPU for training and inference.
In this environment, pure training takes approximately 9 ms per instance
and inference, which also means backtranslation, takes approximately 21 ms per instance.
This means that unsupervised learning approximately needs 30 ms per instance.
WebNLG models use 10.6 million parameters, VG models have 60.7 million parameters.
The difference is due to a larger vocabulary size of 70,800 for VG compared to 8,171 for WebNLG.

\section{Results of all iterations on WebNLG}
See \cref{tab:webnlg-model-select} for all intermediate \gtt{} results
of unsupervised training on WebNLG
and \cref{tab:webnlg-model-select-ie} for \ttg{}.
We find similar trends as for VG (\cref{tab:model-select,tab:unsup:text2graph})
except for \unsup{} being a less reliable performance indicator
for \ttg{} in the sampled noise setting.

\begin{table}[!ht]
	\footnotesize
	\centering
	\setlength\tabcolsep{.5em}
	\begin{tabular}{r|rrr|rrr}
		\toprule
		\multicolumn{1}{c}{}& \multicolumn{3}{c}{sampled noise} & \multicolumn{3}{c}{composed noise}\\
		\cmidrule(lr){2-4}\cmidrule(lr){5-7}
		\multicolumn{1}{c}{\#{}}& \unsup & \val & val & \unsup & \val & val \\
		\midrule
		1 &91.7&12.8&13.0&23.0&15.9&15.5 \\
		2 &\textbf{94.0}&14.7&15.8&53.2&22.2&20.7 \\
		3 &85.2&25.5&26.0&71.0&23.2&22.8 \\
		4 &65.9&27.7&28.8&75.2&25.3&26.2 \\
		5 &65.5&31.4&30.7&69.2&25.9&27.2 \\
		6 &58.1&31.5&31.0&71.5&27.6&27.7 \\
		7 &48.0&31.3&32.3&\textbf{79.2}&29.0&27.7 \\
		8 &48.3&32.8&33.4&52.5&28.1&27.5 \\
		9 &37.5&33.2&34.0&57.1&30.5&30.0 \\
		10 &42.1&32.8&33.4&52.4&30.6&29.9 \\
		11 &38.7&34.7&34.8&59.9&32.0&\textbf{31.6} \\
		12 &38.7&36.4&36.2&42.1&30.4&30.8 \\
		13 &39.3&33.5&35.1&50.0&30.7&30.7 \\
		14 &40.5&36.9&36.6&46.7&30.9&30.7 \\
		15 &41.8&36.5&37.5&48.2&31.1&30.3 \\
		16 &43.2&36.9&38.0&43.7&30.3&29.6 \\
		17 &39.1&35.6&36.6&43.1&29.0&29.7 \\
		18 &38.5&37.5&38.3&31.1&29.7&29.8 \\
		19 &38.8&37.8&38.4&39.5&29.0&29.8 \\
		20 &37.5&37.2&38.6&36.2&31.3&29.8 \\
		21 &36.4&36.8&38.4&35.2&30.0&30.8 \\
		22 &44.8&36.3&39.7&37.6&32.4&30.7 \\
		23 &40.8&35.8&38.2&39.6&31.4&30.3 \\
		24 &35.8&39.2&39.6&39.6&32.4&30.3 \\
		25 &40.6&38.5&39.5&37.0&33.2&30.9 \\
		26 &36.8&38.9&40.3&41.3&32.3&30.2 \\
		27 &44.1&39.7&\textbf{40.6}&37.3&33.0&30.4 \\
		28 &39.3&36.9&38.9&39.0&\textbf{34.7}&30.8 \\
		29 &36.1&37.6&38.6&41.5&31.0&30.6 \\
		30 &38.7&\textbf{40.7}&39.1&42.9&30.6&30.0 \\
		\bottomrule
	\end{tabular}
	\caption{BLEU scores on WebNLG for our unsupervised models evaluated for \gtt{} at different
		iterations. \unsup{} is calculated on all unlabeled data used for training.
		\val{} is a 100-size random sample from val. All results are computed with scripts from \citep{lin18}.}
	\label{tab:webnlg-model-select}
\end{table}

\begin{table}[!ht]
	\footnotesize
	\centering
	\setlength\tabcolsep{.5em}
	\begin{tabular}{r|rrr|rrr}
		\toprule
		\multicolumn{1}{c}{}& \multicolumn{3}{c}{sampled noise} & \multicolumn{3}{c}{composed noise}\\
		\cmidrule(lr){2-4}\cmidrule(lr){5-7}
		\multicolumn{1}{c}{\#{}}& \unsup & \val & val & \unsup & \val & val \\
		\midrule
		1 &\textbf{69.4}&0.0&0.0&0.0&0.0&0.0 \\
		2 &64.0&0.0&0.1&16.2&1.2&1.6 \\
		3 &35.6&0.9&0.3&7.5&3.3&3.0 \\
		4 &47.8&2.6&2.3&37.5&5.5&5.5 \\
		5 &39.2&5.7&3.4&35.3&7.0&6.6 \\
		6 &39.2&6.2&5.6&44.9&9.7&8.0 \\
		7 &45.8&9.8&7.9&58.3&8.0&10.3 \\
		8 &50.0&12.6&10.0&51.1&14.0&12.8 \\
		9 &54.9&13.6&12.9&53.1&12.5&14.0 \\
		10 &58.3&14.9&14.3&51.1&15.9&16.8 \\
		11 &62.5&19.3&17.8&53.8&15.6&17.3 \\
		12 &54.2&20.3&18.2&58.3&16.7&18.0 \\
		13 &57.1&23.1&20.2&47.8&19.8&20.6 \\
		14 &37.5&25.5&21.4&49.0&20.6&22.1 \\
		15 &48.0&25.7&22.4&54.2&23.0&22.8 \\
		16 &52.0&27.9&24.3&46.2&22.5&25.4 \\
		17 &50.0&26.7&25.1&35.6&26.8&26.8 \\
		18 &48.0&32.1&27.7&52.2&27.8&27.7 \\
		19 &56.0&32.3&28.9&58.3&26.4&28.1 \\
		20 &60.0&31.0&30.1&55.3&26.4&29.2 \\
		21 &51.0&32.3&30.4&59.3&27.6&30.7 \\
		22 &55.3&34.9&32.0&\textbf{62.5}&31.7&32.0 \\
		23 &44.9&34.3&32.7&54.9&34.0&32.6 \\
		24 &58.8&38.4&33.7&61.2&31.5&32.4 \\
		25 &46.8&39.6&34.1&58.3&33.3&33.1 \\
		26 &53.8&40.6&36.3&54.2&\textbf{34.4}&32.5 \\
		27 &62.5&41.8&36.4&50.0&33.9&33.3 \\
		28 &55.3&41.0&37.4&40.8&32.6&\textbf{33.7} \\
		29 &56.0&40.7&37.0&58.8&29.5&\textbf{33.7} \\
		30 &59.6&\textbf{41.9}&\textbf{38.5}&53.8&31.6&33.4 \\
		\bottomrule
	\end{tabular}
	\caption{F1 scores on WebNLG for our unsupervised models evaluated for \ttg{} at different
		iterations. \unsup{} is calculated on all unlabeled data used for training.
		\val{} is a 100-size random sample from val.}
	\label{tab:webnlg-model-select-ie}
\end{table}

\end{document}